\documentclass{article}
\usepackage{spconf,amsmath,graphicx,multirow,booktabs,boldline}

\title{MIT-QCRI ARABIC DIALECT IDENTIFICATION SYSTEM 
\\FOR THE 2017 MULTI-GENRE BROADCAST CHALLENGE
}
\name{Suwon Shon$^1$, Ahmed Ali$^2$, James Glass$^1$}
\address{MIT Computer Science and Artificial Intelligence Laboratory (CSAIL), Cambridge, MA, USA$^1$\\
Qatar Computing Research Institute, HBKU, Doha, Qatar$^2$
}
\begin{document}
%
\maketitle
\begin{abstract}

In order to successfully annotate the Arabic speech content found in open-domain media broadcasts, it is essential to be able to process a diverse set of Arabic dialects.  For the 2017 Multi-Genre Broadcast challenge (MGB-3) there were two possible tasks: Arabic speech recognition, and Arabic Dialect Identification (ADI).  In this paper, we describe our efforts to create an ADI system for the MGB-3 challenge, with the goal of distinguishing amongst four major Arabic dialects, as well as Modern Standard Arabic. Our research focused on dialect variability and domain mismatches between the training and test domain. In order to achieve a robust ADI system, we explored both Siamese neural network models to learn similarity and dissimilarities among Arabic dialects, as well as i-vector post-processing to adapt domain mismatches. Both Acoustic and linguistic features were used for the final MGB-3 submissions, with the best primary system achieving 75\% accuracy on the official 10hr test set.

\end{abstract}
\begin{keywords}
Dialect Recognition, Arabic, MGB challenge, Siamese Network, Domain Adaptation
\end{keywords}
\section{Introduction}
\label{sec:intro}

One of the challenges of processing real-world spoken content, such as media broadcasts, is the potential presence of different dialects of a language in the material.  Dialect identification can be a useful capability to identify which dialect is being spoken during a recording.  Dialect identification can be regarded as a special case of language recognition, requiring an ability to discriminate between different members within the same language family, as opposed to across language families (i.e., for language recognition).  The dominant approach, based on i-vector extraction, has proven to be very effective for both language and speaker recognition~\cite{Dehak2011}.  Recently, phonetically aware deep neural models have also been found to be effective in combination with i-vectors~\cite{Richardson2015,Lei2014,Snyder2016}.  Phonetically aware models could be beneficial for dialect identification, since they provide a mechanism to focus attention on small phonetic differences between dialects with predominantly common phonetic inventories.

Since 2015, the Arabic Multi-Genre Broadcast (MGB) Challenge tasks have provided a valuable resource for researchers interested in processing multi-dialectal Arabic speech.  For the ASRU 2017 MGB-3 Challenge, there were two possible tasks.  The first task was aimed at developing an automatic speech recognition system for Arabic dialectal speech based on a multi-genre broadcast audio dataset. The second task was aimed at developing  an Arabic Dialect Identification (ADI) capability for five major Arabic dialects.  This paper reports our experimentation efforts for the ADI task.


While the MGB-3 Arabic ASR task included seven different genres from the broadcast domain, the ADI task focused solely on broadcast news.  Participants were provided high-quality Aljazeera news broadcasts as well as transcriptions generated by a multi-dialect ASR system created from the MGB-2 dataset~\cite{Khurana2017}.  The biggest difference from previous MGB challenges is that only a relatively small development set of in-domain data is provided for adaptation to the test set (i.e., the training data is mismatched with the test data).  For the ADI baseline, participants were also provided with i-vector features from the audio dataset, and lexical features from the transcripts. Evaluation software was shared with all participants using baseline features available via Github\footnote{https://github.com/qcri/dialectID}.

The evaluation scenario for the MGB-3 ADI task can be viewed as channel and domain mismatch because the recording environment of the training data is different from the development and test data.  In general, channel or domain mismatch between training and test data can be a significant factor affecting system performance. Differences in channel, genre, language, topic etc.~produce shifts in low-dimensional projections of the corresponding speech and ultimately cause performance degradations on evaluation data.

In order to address performance degradation of speaker and language recognition systems due to domain mismatches, researchers have proposed various approaches to compensate for, and to adapt to the mismatch \cite{Shum2014,Rahman2015,Singer2015,Garcia-Romero2014,Garcia-Romero2014a,Garcia-Romero2014b,Glembek2014,Aronowitz2014,Shon2017,Shon2017a_recursive,Shon2017b,Kanagasundaram2015,Aronowitz2014a}. For the MGB-3 ADI task, we utilized the development data to adapt to the test data recording domain, and investigated approaches to improve ADI performance both on the domain mismatched scenario, and the matching scenario, by using a recursive whitening transformation, a weighted dialect i-vector model, and a Siamese Neural Network.

In contrast to the language recognition scenario, where there are different linguistic units across languages, language dialects typically share a common phonetic inventory and written language. Thus, we can potentially use ASR outputs such as phones, characters, and lexicons as features.  N-gram histograms of phonemes, characters and lexicons can be used as feature vectors directly, and indeed, a lexicon-based n-gram feature vector was provided for the MGB-3 ADI baseline. The linguistic feature space is, naturally, completely different to the audio feature space, so a fusion of the results from both feature representations has been previously shown to be beneficial~\cite{Ali2016,Hanani2017,Ionescu2017,Rama2017,Malmasi2017}.  Moreover, the linguistic feature has an advantage in channel domain mismatch situations because the transcription itself does not reflect the recording environment, and only contains linguistic information.\footnote{Of course, the word error rate might be higher due to the acoustic mismatch, which could indirectly affect the performance for the linguistic features.}

In this paper, we describe our work for the MGB-3 ADI Challenge.  The final MIT-QCRI submitted system is a combination of audio and linguistic feature-based systems, and includes multiple approaches to address the challenging mismatched conditions. From the official results, this system achieved the best performance among all participants.  The following sections describe our research in greater detail.


\begin{table}[b]
\centering
\resizebox{\linewidth}{!}{%
\begin{tabular}{|cV{2}c|c|c|}
\hline
\begin{tabular}[c]{@{}c@{}}Dataset\\ category\end{tabular}                      & \begin{tabular}[c]{@{}c@{}}Training\\ (TRN)\end{tabular}       & \begin{tabular}[c]{@{}c@{}}Development\\ (DEV)\end{tabular}   & \begin{tabular}[c]{@{}c@{}}Test\\ (TST)\end{tabular}  \\ \hlineB{2}
Size & 53.6 hrs & 10 hrs & 10.1 hrs\\
\hline
Genre                                                                           & \multicolumn{3}{c|}{News Broadcasts}                                                                                                                                                   \\ \hline
\begin{tabular}[c]{@{}c@{}}Channel\\ (recording)\end{tabular}                   & \begin{tabular}[c]{@{}c@{}}Carried out\\ at 16kHz\end{tabular} & \multicolumn{2}{c|}{\begin{tabular}[c]{@{}c@{}}Downloaded directly from\\ a high-quality video server\end{tabular}} \\ \hline

\begin{tabular}[c]{@{}c@{}}Availability\\ for system\\ development\end{tabular} & O                                                              & O                                                             & X                                                     \\ 
\hline
\end{tabular}%
}
\caption{MGB-3 ADI Dataset Properties.}
\label{tab:data}
\end{table}

\section{MGB-3 Arabic Dialect Identification}
\label{sec:format}


For the MGB-3 ADI task, the challenge organizers provided 13,825 utterances (53.6 hours) for the training (TRN) set, 1,524 utterances (10 hours) for a development (DEV) set, and 1,492 utterances (10.1 hours) for a test (TST) set. Each dataset consisted of five Arabic dialects: Egyptian (EGY), Levantine (LEV), Gulf (GLF), North African (NOR), and Modern Standard Arabic (MSA). Detailed statistics of the ADI dataset can be found in~\cite{Ali2017}.  Table~\ref{tab:data} shows some facts about the evaluation conditions and data properties. Note that the development set is relatively small compared to the training set.  However, it is matched with the test set channel domain.  Thus, the development set provides valuable information to adapt or compensate the channel (recording) domain mismatch between the train and test sets.

\section{Dialect Identification Task \& System}
\label{sec:pagestyle}

The MGB-3 ADI task asks participants to classify speech as one of five dialects, by specifying one dialect for each audio file for their submission. Performance is evaluated via three indices: overall accuracy, average precision, and average recall for the five dialects. 

\subsection{Baseline ADI System}
\label{ssec:subhead0}
The challenge organizers provided features and code for a baseline ADI system. The features consisted of 400 dimensional i-vector features for each audio file (based on bottleneck feature inputs for their frame-level acoustic representation), as well as lexical features using bigrams generated from transcriptions~\cite{Ali2017}. For baseline dialect identification, a multi-class Support Vector Machine (SVM) was used.  The baseline i-vector performance was 57.3\%, 60.8\%, and 58.0\% for accuracy, precision and recall respectively.   Lexical features achieved 48.4\%, 51.0\%, and 49.3\%, respectively.  While the audio-based features achieved better performance than the lexical features, both systems only obtained approximately 50\% accuracy, indicating that this ADI task is difficult, considering that there are only five classes to choose from. 

\begin{figure}[b]
\begin{minipage}[b]{.48\linewidth}
  \centering
  \centerline{\includegraphics[width=4.0cm]{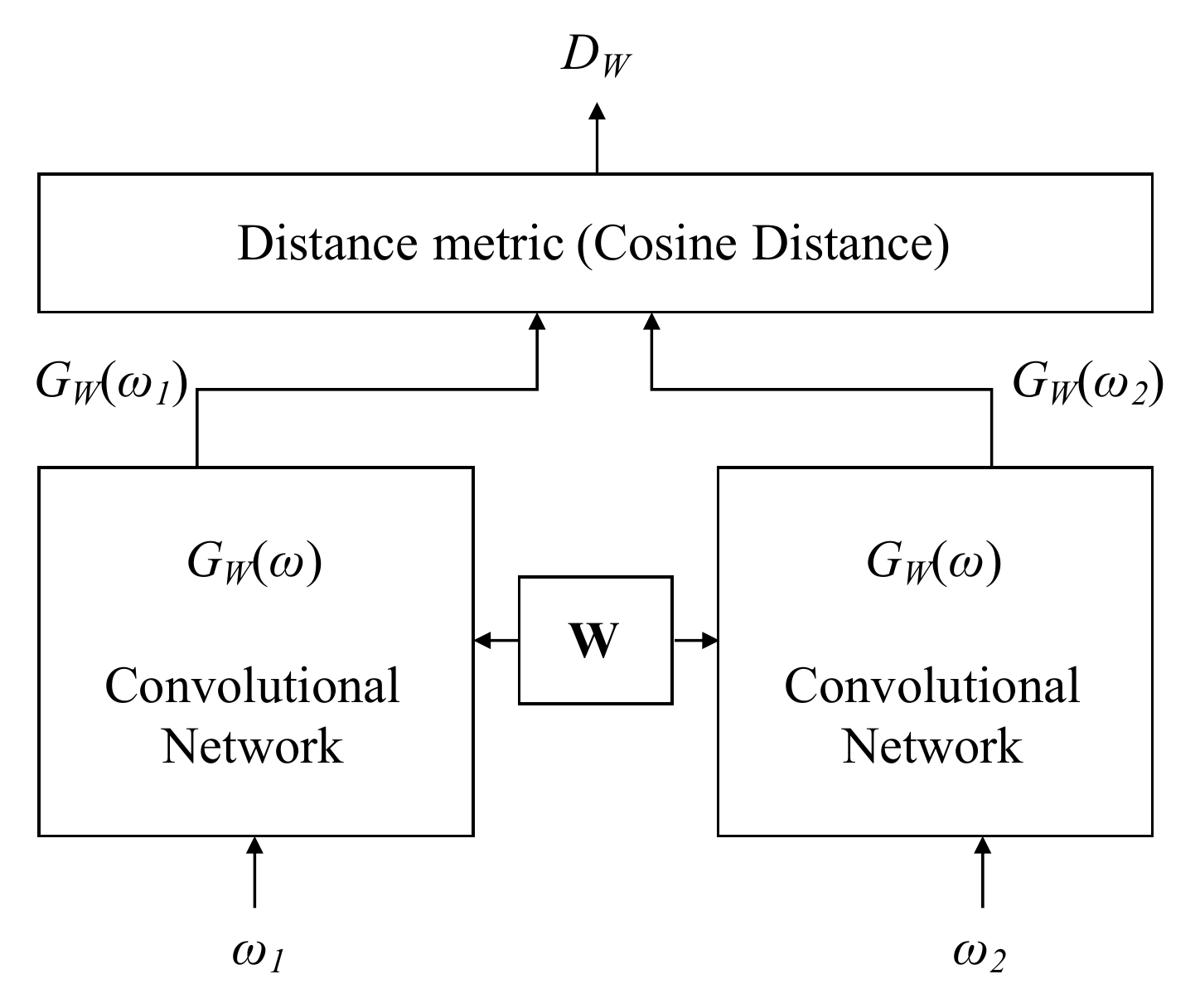}}
  \centerline{(a)}\medskip
\end{minipage}
\hfill
\begin{minipage}[b]{0.48\linewidth}
  \centering
  \centerline{\includegraphics[width=3.0cm]{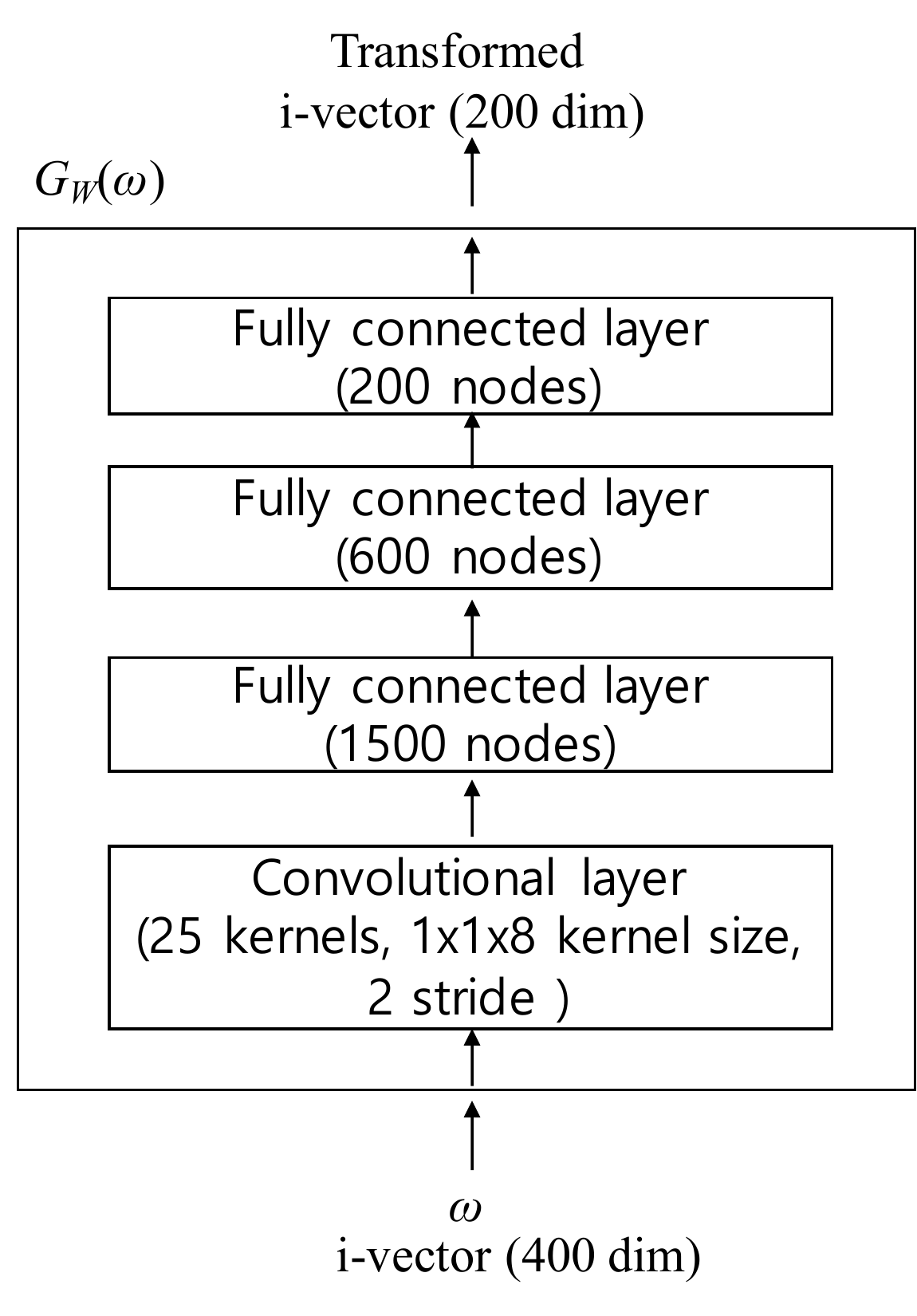}}
  \centerline{(b)}\medskip
\end{minipage}
\caption{(a) Siamese network for i-vector (b) Architecture of convolutional neural network $G_W$}
\label{fig:siam}
\end{figure}

\subsection{Siamese Neural Network-based ADI}
\label{ssec:subhead1}
To further distinguish speech from different Arabic dialects, while making speech from the same dialect more similar, we adopted a Siamese neural network architecture~\cite{Bromley1993} based on an i-vector feature space.  The Siamese neural network has two parallel convolutional networks, $G_W$, that share the same set of weights, $W$, as shown in Figure~\ref{fig:siam}(a). Let $\omega_1$and $\omega_2$ be a pair of i-vectors for which we wish to compute a distance. Let $Y$ be the label for the pair, where $Y$ = 1 if the i-vectors $\omega_1$and $\omega_2$ belong to same dialect, and $Y=-1$ otherwise. To optimize the network, we use a Euclidean distance loss function between the label and the cosine distance, $D_W$, where
\[ L(\omega_i,\omega_j,Y_{ij} = || Y_{ij} - D_W(\omega_i,\omega_j) ||_2^2 \]
For training, i-vector pairs and their corresponding labels can be processed by combinations of i-vectors from the training dataset.  The trained convolutional network $G_W$ transforms an i-vector $\omega$ to a low-dimensional subspace that is more robust for distinguishing dialects.  A detailed illustration of the convolutional network $G_W$ is shown in Figure~\ref{fig:siam}(b).  The final transformed i-vector, $G_W(\omega)$, is a 200-dimensional vector. No nonlinear activation function was used on the fully connected layer. A cosine distance is used for scoring.  

\begin{figure}[t]
\centerline{\includegraphics[width=8.0cm]{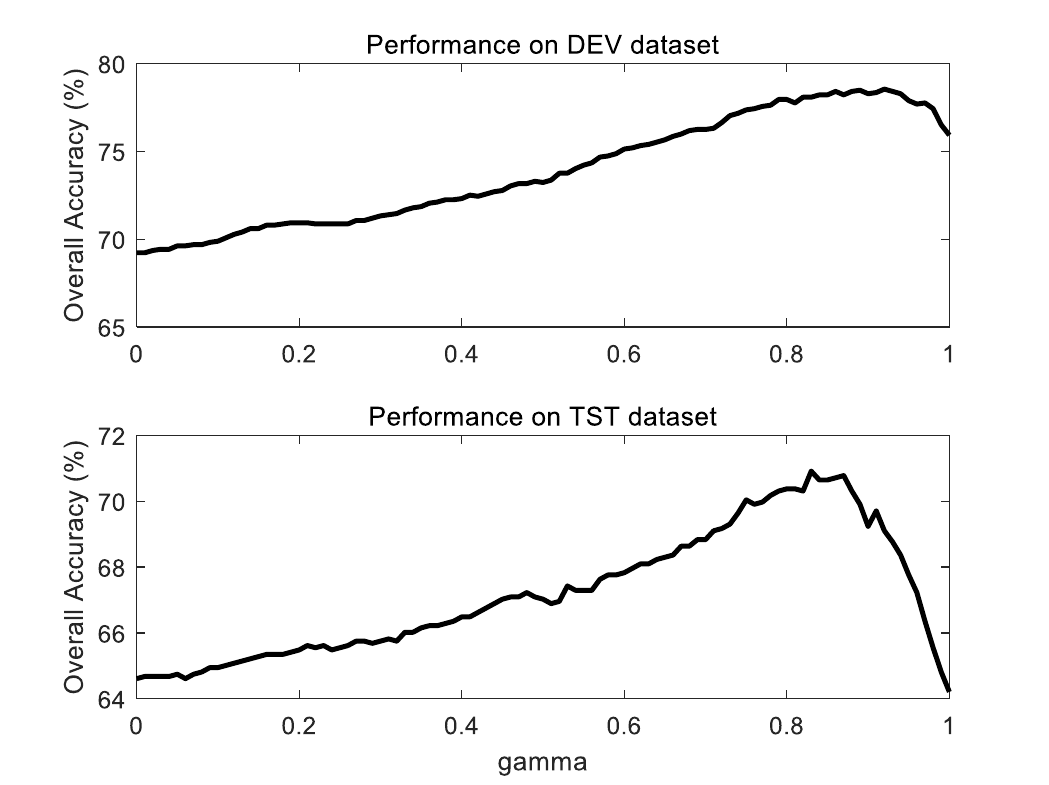}}
\caption{Overall accuracy on DEV and TST sets by gamma: The DEV set shows the best performance at gamma = 0.91, while the TST set shows the best result at gamma=0.83. For our experiments, we used gamma= 0.91.}
\label{fig:interpolated}
\end{figure}

\subsection{i-vector Post-Processing}
\label{ssec:subhead2}
In this section we describe the domain adaptation techniques we investigated using the development set to help adapt our models to the test set.

\subsubsection{Interpolated i-vector Dialect Model}

Although the baseline system used an SVM classifier, Cosine Distance Scoring (CDS) is a fast, simple, and effective method to measure the similarity between an enrolled i-vector dialect model, and a test utterance i-vector. Under CDS, ZT-norm or S-norm can be also applied for score normalization~\cite{Shum2010}. 
Dialect enrollment can be obtained by means of i-vectors for each dialect, and is called the i-vector dialect model: $\overline{\omega_d}=(1/n_d)\sum_{i=1}^{n_d}\omega^d_i$, where $n_d$ is the number of utterances for each dialect $d$.
Since we have two datasets for dialect enrollment, $\overline{\omega^{\text{TRN}}_d}$ for the training set, and $\overline{\omega^{\text{DEV}}_d}$ for the development set, we use an interpolation approach with parameter $\gamma$, where
\[\overline{\omega^{\text{Inter}}_d}=(1-\gamma) \overline{\omega^{\text{TRN}}_d} + \gamma\overline{\omega^{\text{DEV}}_d}\]
We observed that the mismatched training set is useful when combined with matched development set. Figure \ref{fig:interpolated} shows the performance evaluation by parameter $\gamma$ on the same experimental conditions of System 2 in Section 4.3.
This approach can be thought of as exactly the same as score fusion for different system.  However, score fusion is usually performed at the system score level, while this approach uses a combination of knowledge of in-domain and out-of-domain i-vectors with a gamma weight on a single system.

\begin{figure}[t]
\centerline{\includegraphics[width=8.0cm]{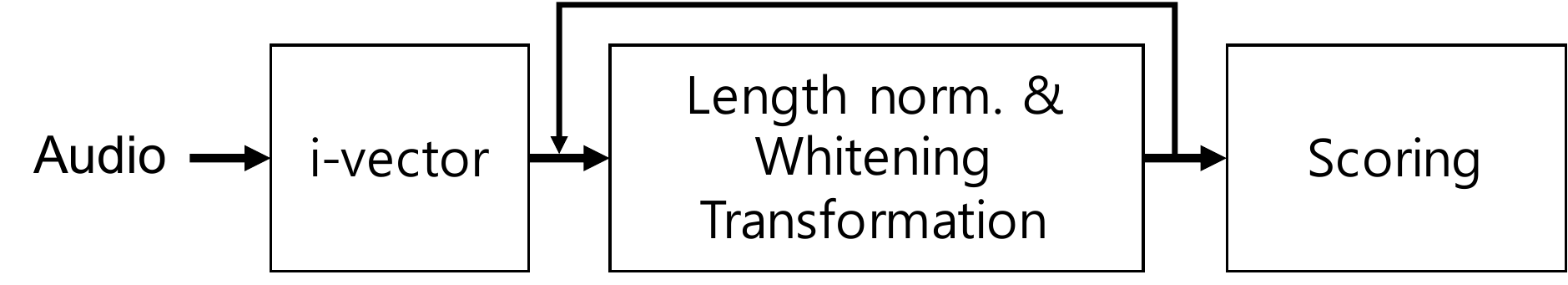}}
\caption{Flowchart of recursive whitening transformation.}
\label{fig:white}
\end{figure}

\subsubsection{Recursive Whitening Transformation}
For i-vector-based speaker and language recognition approaches, a whitening transformation and length normalization is considered essential~\cite{Garcia-Romero2011}. Since length normalization is inherently a nonlinear, non-whitening operation, recently, a recursive whitening transformation has been proposed to reduce residual un-whitened components in the i-vector space, as illustrated in Figure~\ref{fig:white}~\cite{Shon2017a_recursive}.  In this approach, the data subset that best matches the test data is used at each iteration to calculate the whitening transformation.  
In our ADI experiments, we applied 1 to 3 levels of recursive whitening transformation using the training and development data.

\subsection{Phoneme Features}
Phoneme feature extraction consists of extracting the phone sequence, and phone duration statistics using four different speech recognizers: Czech, Hungarian, and Russian using narrowband model, and English using a broadband model~\cite{schwarz2006hierarchical}.  We evaluated the four systems using a Support Vector Machine (SVM). The hyper-parameters for the SVM are distance from the hyperplane (\textit{C} is 0.01), and penalty \textit{l2}. We used the training data for training the SVM and the development data for testing. Table~\ref{tab:phoneme_reco} shows the results for the four phoneme recognizers. The Hungarian phoneme recognition obtained the best results, so we used it for the final system combination.

\begin{table}[ht!]
\centering
\resizebox{\linewidth}{!}{%
\begin{tabular}{|cV{2}c|c|c|}
 \hline
 System  & Accuracy(\%) &Precision(\%)  & Recall(\%)  \\
 \hline
 Czech  & 45 & 45.2  & 45.8  \\
 Hungarian & \textbf{47}& \textbf{47.3} & \textbf{48.1} \\
 Russian  & 46 & 47 & 46.8  \\
 English &  33.3 & 33 & 34 \\
 \hline
\end{tabular}
}
\caption{Evaluating four phoneme recognition systems.}
\label{tab:phoneme_reco}
\end{table}

\begin{table}[ht]
\centering
\resizebox{\linewidth}{!}{%
\begin{tabular}{|cV{2}c|c|c|}
\hline
System (scoring method)   & Accuracy (\%) & Precision (\%) & Recall (\%) \\ \hlineB{2}
i-vector (SVM) - baseline & 57.20         & 60.80          & 58.00       \\ 
i-vector (CDS)            & 56.36         & 59.86          & 57.19       \\ 
LDA i-vector (SVM)        & 60.17         & 60.65          & 60.91       \\ 
LDA i-vector (CDS)        & 58.46         & 61.31          & 59.14       \\ 
Siam i-vector (SVM)       & 61.15         & 62.91          & 61.56       \\ 
Siam i-vector (CDS)       & \textbf{63.65}         & \textbf{64.00}          & \textbf{63.88}       \\ \hline
\end{tabular}%
}
\caption{i-vector evaluation on DEV set: only TRN set is used for training. Note that scores in this table were not calibrated.}
\label{tab:exp_trn}
\end{table}

\begin{table}[ht]
\centering
\resizebox{\linewidth}{!}{%
\begin{tabular}{|cV{2}c|c|c|}
\hline
System (scoring method) & Accuracy (\%) & Precision (\%) & Recall (\%) \\ \hlineB{2}
Baseline word(SVM)      & 48.43         & 50.99          & 49.25       \\ 
Character               & \textbf{57.28}         & \textbf{60.83}          & \textbf{58.03}       \\ 
Phoneme                 & 47.18         & 47.66          & 48.23       \\ 
\hline
\end{tabular}%
}
\caption{Linguistic feature evaluation on DEV set: TRN and DEV sets were used for training.}
\label{tab:exp_lexical}
\end{table}

\begin{table*}[ht!]
\centering
\resizebox{0.7\textwidth}{!}{%
\begin{tabular}{|c|c|cV{2}c|c|c|}
\hline
\begin{tabular}[c]{@{}c@{}}System\\ (scoring method)\end{tabular}              & Whitening               & i-vector Dialect Model  & Accuracy (\%) & Precision (\%) & Recall (\%) \\ \hlineB{2}
i-vector (SVM)                                                                 & \multicolumn{1}{l|}{-} & \multicolumn{1}{lV{2}}{-} & 64.79         & 65.40          & 65.25       \\ \hline
\multirow{4}{*}{\begin{tabular}[c]{@{}c@{}}i-vector \\ (CDS)\end{tabular}}     & Single                 & Averaged               & 68.11         & 68.56          & 68.30       \\ 
                                                                               & Single                 & Interpolated           & 75.52         & 75.87          & 75.66       \\ 
                                                                               & Recursive              & Averaged               & 69.23         & 69.61          & 69.38       \\ 
                                                                               & Recursive              & Interpolated           & \textbf{78.54}         & \textbf{78.75}          & \textbf{78.70}       \\ \hline
\multirow{4}{*}{\begin{tabular}[c]{@{}c@{}}LDA i-vector \\ (CDS)\end{tabular}} & Single                 & Averaged               & 69.16         & 69.50          & 69.35       \\ 
                                                                               & Single                 & Interpolated           & \textbf{69.62}         & \textbf{69.74}          & \textbf{69.82}       \\ 
                                                                               & Recursive              & Averaged               & 68.11         & 68.06          & 68.44       \\ 
                                                                               & Recursive              & Interpolated           & 68.64         & 68.28          & 68.96       \\ \hline
\multirow{4}{*}{\begin{tabular}[c]{@{}c@{}}Siam i-vector\\ (CDS)\end{tabular}} & Single                 & Averaged               & 67.78         & 37.97          & 68.33       \\ 
                                                                               & Single                 & Interpolated           & 76.05         & 76.15          & 76.35       \\ 
                                                                               & Recursive              & Averaged               & 67.65         & 68.18          & 67.86       \\ 
                                                                               & Recursive              & Interpolated           & \textbf{76.31}         & \textbf{76.39}          & \textbf{76.60}    \\           \hline
\end{tabular}%
}
\caption{i-vector evaluation on DEV set: both TRN and DEV sets were used for training.}
\label{tab:exp_trndev}
\end{table*}

\subsection{Character Features}
Word sequences are extracted using a state-of-the-art Arabic speech-to-text transcription system built as part of the MGB-2~\cite{ali2016mgb}. The system is a combination of a Time Delayed Neural Network (TDNN), a Long Short-Term Memory Recurrent Neural Network (LSTM) and Bidirectional LSTM acoustic models, followed by 4-gram and Recurrent Neural Network (RNN) language model rescoring.  Our system uses a grapheme lexicon during both training and decoding. The acoustic models are trained on 1,200 hours of Arabic broadcast speech. We also perform data augmentation (speed and volume perturbation) which gives us three times the original training data. For more details see the system description paper~\cite{Khurana2017}. We kept the \textless UNK\textgreater\  from the ASR system, which indicates out-of-vocabulary (OOV) words, we replaced it with special symbol. Space was inserted between all characters including the word boundaries. An SVM classifier was trained similarly to the one used for the phoneme ASR systems, and we achieved 52\% accuracy,  51.2\% precision and 51.8\% recall. The confusion matrix is different between the phoneme classifier and the character classifier systems, which motivates us to use both of them in the final system combination.

\subsection{Score Calibration}
\label{ssec:subhead3}
All scores are calibrated to be between 0 and 1. A linear calibration is done by the Bosaris toolkit~\cite{Brummer2011}.  Fusion is also done in a linear manner. 

\section{ADI Experiments}
\label{sec:exp}
For experiments and evaluation, we use i-vectors and transcriptions that are provided by the challenge organizers. Please refer to \cite{Ali2017} for descriptions of i-vector extraction and Arabic speech-to-text configuration.

\subsection{Using Training Data for Training}

The first experiment we conducted used only the training data for developing the ADI system. Thus, the interpolated i-vector dialect model cannot be used for this experimental condition. Table~\ref{tab:exp_trn} shows the performance on dimension reduced i-vectors using the Siamese network (Siam i-vector), and Linear Discriminant Analysis (LDA i-vector), as compared to the baseline i-vector system. LDA reduces the 400-dimension i-vector to 4, while the Siamese network reduces it from 400 to 200. Since the Siamese network used a cosine distance for the loss function, the Siam i-vector showed better performance with the CDS scoring method, while others achieved better performance with an SVM. The best system using Siam i-vector showed overall 10\% better performance accuracy, as compared to the baseline.

\subsection{Using Training and Development Data for Training}

For our second experiment, both the training and development data were used for training. For phoneme and character features, we show development set experimental results in Table~\ref{tab:exp_lexical}.  For i-vector experiments, we show results in Table~\ref{tab:exp_trndev}.  In the table we see that the interpolated dialect model gave significant improvements in all three metrics. The recursive whitening transformation gave slight improvements on the original i-vector, but not after LDA and the Siamese network. The best system is the original i-vector with recursive whitening, and an interpolated i-vector dialect model, which achieves over 20\% accuracy improvement over the baseline.

While the Siamese i-vector network helped in the training data only experiments, it does not show any advantage over the baseline i-vector for this condition. We suspect this result is due to the composition of the data used for training the Siamese network. To train the network, i-vector pairs are chosen from from training dataset.  We selected the pairs using both the training and development datasets. However, if we could put more emphasis on the development data, we suspect the Siamese i-vector network would be more robust on the test data. We plan to further examine the performances due to different compositions of data in the future.

\begin{table*}[ht!]
\centering
\resizebox{\textwidth}{!}{%
\begin{tabular}{|c|lV{2}c|c|cV{2}c|c|c|}
\hline
\multicolumn{2}{|cV{2}}{\multirow{2}{*}{\begin{tabular}[|c]{@{}c@{}}System - Only TRN is used\\ (scoring method)\end{tabular}}} & \multicolumn{3}{cV{2}}{DEV}                     & \multicolumn{3}{c|}{TST}                    \\ \cline{3-8} 
\multicolumn{2}{|cV{2}}{}                                                                                                      & Accuracy (\%) & Precision (\%) & Recall (\%) & Accuracy (\%) & Precision (\%) & Recall(\%) \\ \hlineB{2}
\multicolumn{2}{|cV{2}}{i-vector (SVM) - baseline}                                                                             & 57.28         & 60.83          & 58.03       & 55.29         & 59.27          & 56.44      \\ \hline
\multirow{3}{*}{System 1}                                & Siamese i-vector (CDS)                                                   & 63.65         & 64.00          & 63.88       & 60.99         & 60.88          & 61.72      \\ 
                                                         & + score calibration w. DEV dataset                               & 64.44         & 64.76          & 64.70       & 60.92         & 60.80          & 61.62      \\ 
                                                         & + fusion w. char/phone feature                                   & \textbf{66.60}         & \textbf{66.49}          & \textbf{66.86}       & \textbf{67.76}         & \textbf{68.00}          & \textbf{67.88}      \\         \hline
\end{tabular}%
}
\caption{Detailed performance evaluation of submitted system: only TRN dataset was used for training.}
\label{tab:submit_trn}
\end{table*}

\begin{table*}[ht!]
\centering
\resizebox{\textwidth}{!}{%
\begin{tabular}{|c|lV{2}c|c|cV{2}c|c|c|}
\hline
\multicolumn{2}{|cV{2}}{\multirow{2}{*}{\begin{tabular}[c]{@{}c@{}}System - TRN+DEV are used\\ (scoring method)\end{tabular}}} & \multicolumn{3}{cV{2}}{DEV}                         & \multicolumn{3}{c|}{TST}                         \\ \cline{3-8} 
\multicolumn{2}{|cV{2}}{}                                                                                                      & Accuracy (\%)  & Precision (\%) & Recall (\%)    & Accuracy (\%)  & Precision (\%) & Recall(\%)     \\ \hlineB{2}
\multicolumn{2}{|cV{2}}{i-vector (SVM) - baseline}                                                                             & 64.79          & 65.40          & 65.25          & 65.82          & 65.80          & 66.35          \\ \hline
\multirow{6}{*}{System 2}                                         & i-vector (CDS)                                          & 62.07          & 62.51          & 62.63          & 60.86          & 61.87          & 61.49          \\ 
                                                                  & + 1st recursive whitening                               & 68.11          & 68.56          & 68.30          & 63.61          & 64.00          & 64.27          \\ 
                                                                  & + interpolated i-vector dialect model                   & 75.52          & 75.98          & 75.66          & 68.23          & 68.95          & 68.56          \\ 
                                                                  & + 2nd recursive whitening                               & 77.89          & 78.15          & 77.98          & 69.91          & 70.28          & 70.24          \\ 
                                                                  & + 3rd recursive whitening                               & 78.54          & 78.75          & 78.70          & 69.97          & 70.37          & 70.37          \\ 
                                                                  & + fusion w. char/phone feature                          & \textbf{76.38} & \textbf{76.33} & \textbf{76.70} & \textbf{75.00} & \textbf{75.46} & \textbf{75.03} \\ \hline
\multicolumn{1}{|l|}{\multirow{6}{*}{System 3}}                   & Siamese i-vector (CDS)                                  & 65.81          & 66.22          & 66.19          & 62.47          & 62.28          & 63.32          \\ 
\multicolumn{1}{|l|}{}                                            & + 1st recursive whitening                               & 67.78          & 68.33          & 67.97          & 63.54          & 63.53          & 64.22          \\ 
\multicolumn{1}{|l|}{}                                            & + interpolated i-vector dialect model                   & 76.05          & 76.15          & 76.35          & 68.23          & 68.75          & 68.63          \\ 
\multicolumn{1}{|l|}{}                                            & + 2nd recursive whitening                               & 76.18          & 76.26          & 76.49          & 68.30          & 68.81          & 68.69          \\ 
\multicolumn{1}{|l|}{}                                            & + 3rd recursive whitening                               & 76.31          & 76.39          & 76.60          & 68.30          & 68.81          & 68.69          \\ 
\multicolumn{1}{|l|}{}                                            & + fusion w. char/phone feature                          & \textbf{73.43} & \textbf{73.18} & \textbf{73.76} & \textbf{72.72} & \textbf{73.02} & \textbf{72.99} \\ \hline
\end{tabular}%
}
\caption{Performance of MGB-3 ADI systems: TRN and DEV sets used for training. All scores calibrated with DEV set.}
\label{tab:submit_trndev}
\end{table*}

\subsection{Performance Evaluation of Submission}
Tables~\ref{tab:submit_trn} and ~\ref{tab:submit_trndev} show detailed performance evaluations of our three submitted systems.  System 1 was trained using only the training data as shown in Table~\ref{tab:submit_trn}. Systems 2 and 3 were trained using both the training and development sets as shown in Table~\ref{tab:submit_trndev}.  We found the best linear fusion weight based on System 1 to prevent over-fitting was 0.7, 0.2 and 0.1 for i-vector, character, and phonetic based scores respectively. We applied the same weights to Systems 2 and 3 for fusion. 

From Table~\ref{tab:submit_trn}, we see that the Siamese network demonstrates its effectiveness on both the development and test sets without using any information of the test domain. The interpolated i-vector dialect model also demonstrates that it reflects test domain information well as shown by Systems 2 and 3 in Table~\ref{tab:submit_trndev}. Although we expected that the linguistic features would not affected by the domain mismatch, character and phoneme features show useful contributions for all systems. We believe the reason for the performance degradation of Systems 2 and 3 after fusion on the development data can be seen in the fusion rule. We applied the fusion rule derived from System 1 which was not optimal for Systems 2 and 3, considering the development set evaluation. By including the development data as part of their training, Systems 2 and 3 are subsequently overfit on the development data, which was why we used the fusion rule of System 1. From the excellent fusion performance on the test data for Systems 2 and 3, we believe that the fusion rule from System 1 prevented an over-fitted result.

\section{Conclusion}
\label{sec:conclusion}
In this paper, we describe the MIT-QCRI ADI system using both audio and linguistic features for the MGB-3 challenge. We studied several approaches to address dialect variability and domain mismatches between the training and test sets.  Without knowledge of the test domain where the system will be applied, i-vector dimensionality reduction using a Siamese network was found to be useful, while an interpolated i-vector dialect model showed effectiveness with relatively small amounts of test domain information from the development data. On both conditions, fusion of audio and linguistic feature guarantees substantial improvements on dialect identification. As these approaches are not limited to dialect identification, we plan to explore their utility on other speaker and language recognition problems in the future.

\bibliographystyle{IEEEbib}
\bibliography{strings,refs}

\end{document}